\documentclass[final,hidelinks,onefignum,onetabnum]{siamart250211}
\usepackage{amsmath,amssymb,amsfonts,graphicx,bm}
\usepackage{enumitem,algorithm,algpseudocode}

\DeclareMathOperator*{\argmin}{arg\,min}
\newsiamthm{assumption}{Assumption}
\newsiamthm{example}{Example}
\newsiamthm{remark}{Remark}

\headers{Learning task-specific predictive models}{Yin and Li}
\title{Learning task-specific predictive models for scientific computing}
\author{
Jianyuan Yin\thanks{Department of Mathematics, National University of Singapore.}
\and
Qianxiao Li\thanks{Department of Mathematics \& Institute for Functional Intelligent Materials, National University
of Singapore(\email{qianxiao@nus.edu.sg}).}
}

\begin{document}

\maketitle
\begin{abstract}
We consider learning a predictive model to be subsequently used for a given downstream task (described by an algorithm) that requires access to the model evaluation. This task need not be prediction, and this situation is frequently encountered in machine-learning-augmented scientific computing. We show that this setting differs from classical supervised learning, and in general it cannot be solved by minimizing the mean square error of the model predictions as is frequently performed in the literature. Instead, we find that the maximum prediction error on the support of the downstream task algorithm can serve as an effective estimate for the subsequent task performance. With this insight, we formulate a task-specific supervised learning problem based on the given sampling measure, whose solution serves as a reliable surrogate model for the downstream task. Then, we discretize the empirical risk based on training data, and develop an iterative algorithm to solve the task-specific supervised learning problem. Three illustrative numerical examples on trajectory prediction, optimal control and minimum energy path computation demonstrate the effectiveness of the approach.
\end{abstract}



\section{Introduction}
The classical supervised learning problem aims to learn an unknown mapping $f^*$
from the input space $\mathcal{X}$ to the output space $\mathcal{Y}$ using
training data $(x, y=f^*(x))$ drawn from a sampling measure $x\sim \rho$.
To do this, we often minimize an empirical estimation of the population risk,
e.g. in the form of the mean square error (MSE),
\begin{equation} \label{eqn:p1}
\min_{\theta\in\Theta} \quad L(\theta) = \underset{x\sim\rho}{\mathbb{E}} \|f_\theta(x) - f^*(x)\|_{\mathcal{Y}}^2,\tag{P1}
\end{equation}
over a hypothesis space $\mathcal{F}=\{f_\theta: \mathcal{X}\to
\mathcal{Y}\;|\;\theta \in\Theta\}$, where $\|\cdot\|_{\mathcal{Y}}$ is a given
norm of $\mathcal{Y}$.  In the sense of expectation, the minimization problem
\eqref{eqn:p1} gives an optimal surrogate model for predicting the output
corresponding to a given input that is drawn from the sampling measure $\rho$.
Besides MSE, the discussion can be generalized to other forms of risk as well.

While this setting is often sufficient for predictive problems such as
regression or classification, many emerging applications of machine
learning---particularly in science and engineering---require a different
criterion for assessing model performance.  For example, in learning dynamical
systems, we train a model on short temporal trajectories, but our goal could be
to estimate the asymptotic behaviors of the system
\cite{yu2021onsagernet,lin2022data}.
Especially, machine-learning-based potentials for molecular simulations are highly valuable, gaining widespread recognition and applications \cite{chmiela2017machine, zhang2018deep, doerr2021torchmd, unke2021machine, zhang2024pretraining}.
Another class of examples are data-driven inverse problems, where the quality of the trained forward model
mapping inputs to outputs is assessed based on its ability to find an input that gives a target output
\cite{aggarwal2018modl,raissi2019physics,karniadakis2021physics},
including control problems as a special case \cite{zhang2022initial, mowlavi2023optimal}.
In all such cases, it is not clear---in fact, not true, as we show later---that an
optimal predictive model solving \eqref{eqn:p1} is also the best model for
accomplishing these downstream tasks.  This prompts us to revisit how such
learning problems should be correctly formulated, how they are related to but
departs from the classical supervised learning problem \eqref{eqn:p1}, and how
these models should be trained to achieve the desired outcomes in applications \cite{shankar2025differentiable,giovanis2024polynomial,de2024ai,chang2025leveraging}.

In this paper, we propose a reformulation of the supervised learning problem that takes into account a given general downstream task.
This downstream task can be abstracted as an algorithm $\mathcal{A}$ that requires access to the model evaluation on a specific subset of the input space.
With this abstraction, we can formulate an alternative task-specific supervised learning problem whose solution serves as a reliable surrogate model for the downstream task algorithm.
Moreover, we discuss how this new supervised learning problem is related to \eqref{eqn:p1}.
To solve this supervised learning problem, we turn to minimize an upper bound, and present an empirical risk based on training data from the sampling measure.
Finally, we give a practical iterative algorithm to solve this new supervised learning problem, and demonstrate its performance on a wide range of illustrative problems involving trajectory prediction of dynamical systems, tracking problems in optimal control, and the problem of finding the minimum energy path on the energy landscape.
Moreover, we conduct ablation studies to show the impact of the input sampling measures and hypothesis space sizes
on the performance of the proposed method.

\section{Problem formulation}
Contrary to classical supervised learning, we consider a situation in which we
train the model $f_\theta \in \mathcal{F}$ to approximate a ground truth $f^*$
for a downstream task known before training, which may be different from prediction.
For example, this downstream task could be specified as
\begin{enumerate}
\item Inverse design problems: for a forward model $f^*: \mathcal{X} \to \mathcal{Y}$,
find $x^*$ to minimize $\|f^*(x^*) - y^*\|_{\mathcal{Y}}^2$ for a given output $y^*\in \mathcal{Y}$.
\item Trajectory prediction from vector fields:
for a vector field $f^*: \mathcal{X} \to \mathcal{X}$,
find the solution to $\dot{x}=f^*(x)$ up to time $T$ from a given initial point $x(0)\in\mathcal{X}$.
\item Trajectory prediction from flow maps: for a flow map $f^*:\mathcal{X} \to \mathcal{X}$ over one time step,
find the trajectory $[f^{*(t)}(x_0)]_{t=1}^T$ from a given initial point $x_0\in \mathcal{X}$.
\item Calculation of the minimum energy path (MEP): for a gradient system $f^*=-\nabla E$ of an energy landscape $E$, find the MEP between two minima of $E$ \cite{weinan2010transition}.
\end{enumerate}
Before training, we assume that a specific algorithm $\mathcal{A}$ is
developed based on scientific computing to solve the downstream task.
The downstream task algorithm $\mathcal{A}$ requires a continuous function $f:\mathcal{X}\to\mathcal{Y}$ as input and gives the output $\mathcal{A}(f)$ in a Banach space $\mathcal{B}$.
Furthermore, the algorithm should exhibit stability; that is, when the input is perturbed slightly, the output error can be controlled by the input error.
We give a detailed description of the algorithm $\mathcal{A}$ with some mild assumptions.
\begin{assumption}\label{ass:1}
The downstream task algorithm $\mathcal{A}$ does not require the analytical form of the model $f$,
but only evaluates $f$ as a black-box function on a finite sequence of points,
denoted as the \emph{algorithm support} $\mathcal{S}(f)=[s_1(f),\cdots,s_{J}(f)]\in \mathcal{X}^{J}$.
Each $s_j(f)$ is a function $S_j:\mathcal{X}^{j-1}\to \mathcal{X}$ of previous evaluations of $f$:
\begin{equation}\label{eqn:sj}
s_1(f)=S_1,\quad
s_j(f) = S_j(f(s_1(f)), \cdots, f(s_{j-1}(f))), \quad j=2,\cdots,J.
\end{equation}
The algorithm output $\mathcal{A}(f) = A\left(f(s_1(f)), \cdots, f(s_{J}(f))\right)$ is a function $A:\mathcal{X}^J\to \mathcal{Y}$ of these evaluations.
We assume that $f^*$, $A$ and $S_j$ are Lipschitz continuous functions.
\end{assumption}
\begin{remark}
Although the algorithm support is assumed to have a fixed length $J$ for simplicity,
our method can be directly applied to variable length cases.
\end{remark}

The following is an example of the downstream task algorithm for illustration.
\begin{example}[Explicit Euler scheme]\label{exm}
Consider learning an autonomous dynamical system $\dot{x}=f^*(x)$ from vector field data $(x, y=f^*(x)) \in \mathbb{R}^d\times \mathbb{R}^d$, while the downstream task is to predict the trajectory $x(t)$ from a given point $x(0) \in \mathbb{R}^d$ over the time interval $t\in[0,T]$.
We assume that $f^*$ is a Lipschitz continuous function with Lipschitz constant $C_L$.
Here we choose a simple algorithm---the explicit Euler scheme with a sufficiently small step size $\tau=T/J$.
We have,
$$\mathcal{A}(f) = [s_2(f), \cdots, s_{J+1}(f)], \text{ where } s_1(f) = x(0), \; s_{j+1}(f) = s_j(f) + \tau f(s_j(f)).$$
\end{example}

Based on training data for $f^*$, a typical data-driven method to solve the downstream task is to train a model $f_\theta\in \mathcal{F}$ and apply $f_\theta$ as a surrogate to the downstream task algorithm $\mathcal{A}$.
The hope is that the algorithm $\mathcal{A}$ with the surrogate model $f_\theta$
gives an output $\mathcal{A}(f_\theta)$ that closely
approximates the algorithm output with the ground truth $\mathcal{A}(f^*)$.
In other words, our goal is to minimize the error of the algorithm output using the surrogate model:
\begin{equation} \label{eqn:p2}
\min_{\theta\in\Theta} \quad J_{\mathcal{A}}(\theta) = \|\mathcal{A}(f_\theta)- \mathcal{A}(f^*)\|_{\mathcal{B}}.\tag{P2}
\end{equation}

\textbf{Relationship between \eqref{eqn:p1} and \eqref{eqn:p2}.}
We show that there is an essential difference between \eqref{eqn:p1} and \eqref{eqn:p2} that is not attributed to generalization error or finite sampling error during empirical risk minimization, but stems from the distribution shift (also known as distribution mismatch or deviation) between the sampling environment in training and the application environment in the downstream task, which has been reported in various scenarios \cite{zhang2019active, zhang2022initial, zhao2025mitigating}.
The surrogate model $f_\theta$ is trained in the sampling environment $\rho$, while the downstream task applies the surrogate model $f_\theta$ in a different environment $\mathcal{S}(f_\theta)$.
We assume that training data can provide enough information for the downstream task.
\begin{assumption}\label{ass:2}
The training data is sampled from a sampling measure $x\sim \rho$ with noiseless labels $y=f^*(x)$,
and we cannot resample new data from a different measure.
Furthermore, the ground truth algorithm support $\mathcal{S}(f^*)$ is included in $\operatorname{supp}\rho$.
To enable the training procedure,
we also assume that the algorithm support of any surrogate model $\mathcal{S}(f_\theta)$
involved in the training process is included in $\operatorname{supp}\rho$.
\end{assumption}

If the ground truth $f^*$ is included in the hypothesis space $\mathcal{F}$ and $\operatorname{supp}\rho=\mathcal{X}$,
the global minimizer in \eqref{eqn:p1} would be exactly $f_\theta = f^*$, so the solution to \eqref{eqn:p1} would also be a solution to \eqref{eqn:p2}.
In this case \eqref{eqn:p2} can be solved by minimizing MSE in \eqref{eqn:p1}.
Although this training procedure could be inefficient, the obtained model can serve as an accurate model for the downstream task.
However, if the ground truth $f^*$ is not exactly included in the hypothesis space $\mathcal{F}$,
it turns out that the global minimizer of \eqref{eqn:p1} would not in general be a minimizer of \eqref{eqn:p2}.
In contrast, the minimizer of \eqref{eqn:p1} may lead to a error $J_{\mathcal{A}}$ much larger than what can be achieved
if \eqref{eqn:p2} were minimized.
We use a specific case of \cref{exm} to illustrate this observation.

\begin{example}\label{exm:2}
Consider a one-dimensional case of \cref{exm} as $f^*(x) = -\alpha |x|$
with $\rho=\mathcal{U}([-1,1])$, $\alpha \tau\in(0,1)$, $x(0)=1$, and adopt a hypothesis space of affine functions,
$\mathcal{F}=\{f_\theta(x)=\theta_0+\theta_1 x \;|\; \theta=(\theta_0, \theta_1)=\mathbb{R}^2\}$.
The minimizer of \eqref{eqn:p1} is $f_{\theta_1}(x) = -\alpha/2$ with $s_j(f_{\theta_1}) = 1-(j-1)\alpha\tau/2$, while the correct trajectory is $s_j(f^*)=(1-\alpha\tau)^{j-1}$.
Therefore, the algorithm output error of $f_{\theta_1}$
\begin{equation}\label{eqn:ex2}
J_{\mathcal{A}}^2(\theta_1) = \|\mathcal{A}(f_{\theta_1}) - \mathcal{A}(f^*)\|_2^2 = \sum_{j=1}^J [1-j\alpha\tau/2 - (1-\alpha\tau)^{j}]^2,
\end{equation}
could be very large.
However, $f_{\theta_2}(x)=-\alpha x$ is a minimizer of $J_{\mathcal{A}}$ with $J_{\mathcal{A}}(\theta_2)=0$.
Therefore, this hypothesis space is, in fact, large enough to deal with \eqref{eqn:p2}, while the failure comes from the bad surrogate model obtained from \eqref{eqn:p1}.
\end{example}

In \cref{exm:2}, calculating $\mathcal{A}(f^*)$ does not require any evaluations of $f^*$ at $x<0$, so it is not required to approximate $f^*$ for $x<0$.
On the other hand, $f^*$ is evaluated on many points in $[0,1]$, so the prediction error of $f_\theta$ in $x\in[0,1]$ should be
properly minimized during training.
Since \eqref{eqn:p1} treats these two parts with equal importance, \eqref{eqn:p2} should be considered as a different problem that cannot be effectively solved from \eqref{eqn:p1}.

Intuitively, solving \eqref{eqn:p1} minimizes MSE between $f^*$ and $f_\theta$
with the sampling measure $\rho$ as effective sample weights (relative to the uniform measure),
while \eqref{eqn:p2} requires a high precision of $f_\theta$ only at part of the input space
$\mathcal{X}$.
In other words, \eqref{eqn:p2} actually involves a distribution other than $\rho$ where the surrogate model should predict well, and the algorithm support $\mathcal{S}(f_\theta)$ can typically represent this undetermined distribution.
We refer to this as a \textit{task distribution} of $f_\theta$ (while using the term ``distribution'' informally),
and the algorithm support $\mathcal{S}(f_\theta)$ is viewed as a finite sample from this distribution.
To implement the algorithm $\mathcal{A}$, we need to train a surrogate model that performs well on this task distribution instead of the sampling measure.
When we consider \eqref{eqn:p2} as the only purpose for training, it is not important to consider the prediction error on inputs far away from $\mathcal{S}(f_\theta)$, so the prediction error of $f_\theta$ on these training data could be ignored and sacrificed during training.
The distribution shift between the sampling measure $\rho$ and
the task distribution is the essential reason why \eqref{eqn:p1} and \eqref{eqn:p2} perform differently.
However, most existing approaches---especially in machine-learning-augmented scientific computing---often
simply use \eqref{eqn:p1} as a surrogate optimization problem for \eqref{eqn:p2}.
In the following we show that better performance can be obtained
by accounting for the task distribution during training.

\section{Methods}
In view of the previous discussions,
to effectively solve the downstream task we should solve \eqref{eqn:p2}.
However, the objective $J_{\mathcal{A}}$ in \eqref{eqn:p2} cannot be
directly minimized due to the inaccessibility of $\mathcal{A}(f^*)$.
To resolve this, we establish an upper bound for the objective $J_{\mathcal{A}}$
that can then be approximated by a sample-reweighted loss function
evaluated on the training data.
From an initial model $f_\theta$ obtained by \cref{eqn:p1}, our training method is to optimize the empirical risk iteratively.
To enable training, we alternatively calculate the sample-reweighting coefficients
on the training samples induced by the surrogate model and implement the subsequent
reweighted empirical risk minimization to obtain an updated surrogate model.
This training process is iterated until a stopping criterion is reached,
and an early stopping strategy is applied to obtain a final surrogate model due to lack of validation data.
The complete algorithm is summarized in \cref{alg}.

\begin{figure}[htbp]
    \centering
    \includegraphics[width=\linewidth]{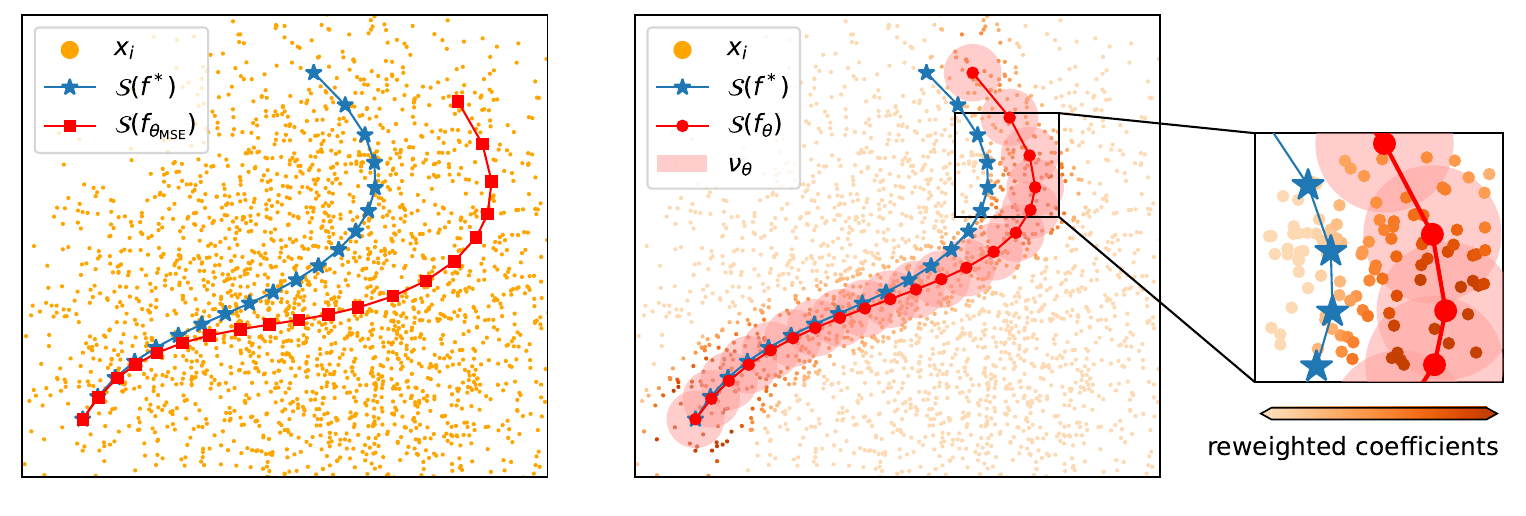}
    \caption{An overview illustration figure of the task-specific supervised learning problem.
    Blue stars: the algorithm support of the ground truth.
    Red squares (left): the algorithm support of the MSE surrogate model $f_{\theta_{\mathrm{MSE}}}$.
    Red dots (right): the algorithm support of the surrogate model $f_\theta$.
    Red regions: task distribution $\nu_\theta$.
    Orange points: training data.
    Color from light to dark (right) represents reweighted coefficients from small to large.
    }
    \label{fig:0}
\end{figure}

Before presenting the details of our method, we provide an overview in \cref{fig:0} to illustrate the task-specific supervised learning problem with a two-dimensional vector field in \cref{exm}.
The orange point cloud represents training data distributed according to $\rho$,
while the blue stars represent the algorithm support of the ground truth $f^*$.
However, the MSE surrogate model $f_{\theta_{\text{MSE}}}$
has a different support (illustrated by red squares).
Therefore, for the current surrogate model $f_\theta$, we apply a task distribution $\nu_\theta$ (illustrated by red regions)
to describe the neighbourhoods of $\mathcal{S}(f_\theta)$ (illustrated by red dots),
and try to improve the accuracy of the current surrogate model $f_\theta$ based on $\nu_\theta$.
To realize this, the task-specific supervised learning problem attaches large reweighting coefficients to
those training data in the red regions,
and the iterative training algorithm updates the surrogate model by minimizing a reweighted error,
which approximates the maximum prediction error on these regions.
Finally, because the surrogate model predicts well on the current support $\mathcal{S}(f_\theta)$,
this support should get closer to $\mathcal{S}(f^*)$ during training,
so that the downstream task algorithm should produce a good output based on the surrogate model.

\subsection{Establishing an upper bound}
Intuitively, if the surrogate model $f_\theta$ gives predictions with small errors throughout the algorithm execution, the algorithm output $\mathcal{A}(f_\theta)$ should be close to the accurate output $\mathcal{A}(f^*)$.
In other words, the prediction error of $f_\theta$ on $\mathcal{S}(f_\theta)$ should be provide control over $J_{\mathcal{A}}(\theta)$.
In \cref{exm}, it can be obtained through direct calculations that
\begin{equation}\label{eqn:ex3}
\|s_j(f_\theta) - s_j(f^*)\| \leq C_L^{-1}((1+\tau C_L)^{j-1}-1) \max_{x \in \mathcal{S}(f_\theta)} \|f_\theta(x) - f^*(x) \|.
\end{equation}
Hence, the algorithm output error can be controlled by the maximum error on the algorithm support $\mathcal{S}(f_\theta)$.
In general, we have the following proposition that bounds the error $J_{\mathcal{A}}$ in \eqref{eqn:p2}
in terms of the maximum prediction error of a model $f$ on the support $\mathcal{S}(f)$.
The proof follows directly from Lipschitz continuity, and explicit estimates on the constants are omitted for simplicity.
\begin{proposition}\label{pro:sup}
Under \cref{ass:1}, from the Lipschitz continuity of $S_j$ and $f^*$, we have, for $j=1,\cdots,J$,
\begin{equation}\label{eqn:pp1}
\|s_j(f) - s_j(f^*)\|_{\mathcal{X}} +
\|f(s_j(f)) - f^*(s_j(f^*))\|_{\mathcal{Y}}
\lesssim \max_{x \in \mathcal{S}(f)} \|f(x) - f^*(x) \|_{\mathcal{Y}}.
\end{equation}
Consequently, from the Lipschitz continuity of $\mathcal{A}$, the algorithm output error $\|\mathcal{A}(f) - \mathcal{A}(f^*)\|_{\mathcal{B}}$
can be directly controlled by $\max_{x \in \mathcal{S}(f)} \|f(x) - f^*(x) \|_{\mathcal{Y}}$ up to a constant.
\end{proposition}
Based on \cref{pro:sup}, in order to control $J_{\mathcal{A}}(\theta)$, we can establish an upper bound $R_{\mathcal{S}}$ for the objective $J_{\mathcal{A}}$ in \eqref{eqn:p2} as:
\begin{equation}\label{eqn:rs}
\min_{\theta\in\Theta} \quad R_{\mathcal{S}}(\theta) = \max_{x\in \mathcal{S}(f_\theta)} \|f_\theta(x) - f^*(x)\|_{\mathcal{Y}}^2,
\end{equation}
which satisfies $J_{\mathcal{A}}^2(\theta)\lesssim R_{\mathcal{S}}(\theta)$.
Here, $R_{\mathcal{S}}$ denotes the maximum prediction error between $f_\theta$ and $f^*$ on the algorithm support $\mathcal{S}(f_\theta)$.
By minimizing $R_{\mathcal{S}}$, the surrogate model $f_\theta$ would be uniformly close to the ground truth at each point of the algorithm support ${S}(f_\theta)$ up to an error bound related to $R_{\mathcal{S}}(\theta)$.
Note that here we employ the maximum square error in accordance with the MSE in \eqref{eqn:p1}.
In general, if \eqref{eqn:p1} employs a different loss function, the error bound holds
with the maximum square loss replaced by the corresponding maximum over the loss employed in \eqref{eqn:p1}.

\subsection{A task-specific supervised learning problem}
Although the objective $R_{\mathcal{S}}$ in \eqref{eqn:rs} gives an upper bound for solving \eqref{eqn:p2}, it is still difficult to optimize directly.
While the algorithm support $\mathcal{S}(f_\theta)$ is accessible by implementing the algorithm, it changes with training due to $\theta$ being updated.
More importantly, we do not have labels $f^*(x)$ for $x\in \mathcal{S}(f_\theta)$ without resampling new data, so $\|f_\theta(x)- f^*(x)\|_{\mathcal{Y}}$ is also unknown throughout the training process.
Therefore, we need to derive another population risk on the sampling measure $\rho$ to approximate $R_{\mathcal{S}}$.

\textbf{Estimation of the maximum error.}
Let us denote all probability measures of $\mathcal{X}$ whose support is $\mathcal{S}(f_\theta)$ by $U_\theta$.
The maximum error $R_{\mathcal{S}}$ can be rewritten as
\begin{equation}\label{eqn:sup}
\begin{aligned}
R_{\mathcal{S}}(\theta)
= \max_{x\in \mathcal{S}(f_\theta)} \|f_\theta(x) - f^*(x)\|_{\mathcal{Y}}^2
= \sup_{\mu\in U_\theta}\underset{x\sim \mu}{\mathbb{E}} \|f_\theta(x) - f^*(x)\|_{\mathcal{Y}}^2.
\end{aligned}
\end{equation}
Although there exist methods that directly minimize a supremum error, for example in adversarial training \cite{tramer2019adversarial,wang2021adversarial}, we instead minimize an approximation of $R_{\mathcal{S}}$ for stability.
Instead of trying to attain the supremum, we take a particular probability measure $\mu_\theta$ to give a moderate approximation.

For each point $x\in\mathcal{S}(f_\theta)$, the surrogate model is expected to be close to the ground truth not only at this point, but also in its neighbourhood.
From the stability of the algorithm $\mathcal{A}$,
the algorithm needs to be robust under small perturbations,
so the surrogate model should also be accurate in these neighbourhoods.
Therefore, we apply kernel density estimation methods with a kernel $\kappa$
instead of the delta measure $\delta_x$ for each point $x$ in the algorithm support.
In the following, a kernel $\kappa(\cdot, x)$ is a smooth probability distribution function
centered on $x\in\mathcal{X}$ with variance $\epsilon^2$,
such that $\kappa(\cdot, x)$ converges to $\delta_x$ weakly as $\epsilon\to 0$.
The kernel density estimation will be applied multiple times in subsequent derivations.

If the algorithm support $\mathcal{S}(f_\theta)$ is distributed unevenly,
an outlier in $\mathcal{S}(f_\theta)$ may not be adequately emphasized.
Therefore, we regard $\mathcal{S}(f_\theta)$ as fixed-size samples drawn from an empirical task distribution,
rather than considering all the points in $\mathcal{S}(f_\theta)$ to be of equal importance.
A stratification factor $\alpha_\theta(x)$ (defined in \eqref{eqn:nutheta})
is introduced to emphasize outliers, and a particular probability measure $\mu_\theta$ is taken as
\begin{equation}\label{eqn:mut}
\frac{\mathrm{d}\mu_\theta}{\mathrm{d}x} = \sum_{x\in \mathcal{S}(f_\theta)} \omega_\theta(x) \alpha_\theta(x)\kappa(\cdot, x),
\end{equation}
to approximate the supremum in \eqref{eqn:sup}.
Here, $\omega_\theta: \mathcal{S}(f_\theta)\to \mathbb{R}_+$ is a weight function to emphasize points with large prediction errors.
Intuitively, a larger weight should be attached to a point in $\mathcal{S}(f_\theta)$ with a larger prediction error accordingly.
As the kernel variance $\epsilon^2$ tends to zero, $\mu_\theta$ weakly converges to a measure in $U_\theta$.
For stability in practical computations, this variance is not taken to be too small.

\textbf{Change of measure.}
From \cref{ass:2}, our training data is sampled from a fixed sampling measure $\rho$, so we should rewrite $R_{\mathcal{S}}$ based on this measure.
By taking $\mu$ in \eqref{eqn:sup} as $\mu_\theta$ \eqref{eqn:mut}, we obtain a population risk as
\begin{equation}\label{eqn:q}
R(\theta) = \underset{x\sim \mu_\theta}{\mathbb{E}} \|f_\theta(x)-f^*(x)\|_{\mathcal{Y}}^2 =
\underset{x\sim \rho}{\mathbb{E}}\left[\frac{\mathrm{d}\mu_\theta}{\mathrm{d}\rho}(x) \|f_\theta(x)- f^*(x)\|_{\mathcal{Y}}^2  \right],
\end{equation}
as an approximation of $R_{\mathcal{S}}(\theta)$.
The measure in the population risk is changed from $\mu_\theta$ to the sampling measure $\rho$ based on the RN theorem.
Assumed that $\kappa(\cdot, x)$ has bounded support and the variance $\epsilon^2$ is small enough, the existence of the RN derivative $\frac{\mathrm{d}\mu_\theta}{\mathrm{d}\rho}(x)$ in \eqref{eqn:q} can be guaranteed by \cref{ass:2}.
As a consequence, we obtain a task-specific population risk on the sampling measure $\rho$ to optimize as,
\begin{equation}\label{eqn:rp}
\begin{aligned}
&\min_{\theta\in\Theta} \quad R(\theta) =
\underset{x\sim \rho}{\mathbb{E}}\left[\frac{\mathrm{d}\mu_\theta}{\mathrm{d}\rho}(x) \|f_\theta(x)- f^*(x)\|_{\mathcal{Y}}^2   \right],\\
&\text{s.t.}\quad
\frac{\mathrm{d}\mu_\theta}{\mathrm{d}x}(x) = \sum_{x\in \mathcal{S}(f_\theta)} \omega_\theta(x) \alpha_\theta(x)\kappa(\cdot, x),
\end{aligned}
\end{equation}
We refer to the problem \eqref{eqn:rp} as a \textit{task-specific supervised learning problem}.

The task-specific population risk $R$ in \cref{eqn:rp} is the core minimization target in our method.
At first glance, the task-specific population risk $R$ looks like a reweighted version of the original MSE in \eqref{eqn:p1}, while it differs significantly from a sample-reweighted loss commonly seen for training \cite{ren2018learning, shen2020stable}, because the reweighting coefficients $\frac{\mathrm{d}\mu_\theta}{\mathrm{d}\rho}(x)$ in \eqref{eqn:rp} are not constants or hyperparameters.
In particular, they cannot be calculated before training, but are determined by both the downstream task and the surrogate model $f_\theta$ that is currently being trained.
As the surrogate model $f_\theta$ is updated, the RN derivative $\frac{\mathrm{d}\mu_\theta}{\mathrm{d}\rho}$ also varies according to the algorithm support.
Therefore, \eqref{eqn:rp} could be highly nonlinear with respect to $\theta$.

If we can sample new training data near the algorithm support $\mathcal{S}(f_\theta)$ at a low cost, i.e. resample from $\mu_\theta$, we can directly minimize \cref{eqn:rs} based on these data instead of reweighting samples in \cref{eqn:rp}.
Some research have reported that resampling methods often outperform reweighting methods \cite{seiffert2008resampling, an2021resampling}.
Here, we only consider cases with fixed sampling measure, which allows for more general application to cases
where new data cannot be easily resampled,
and we leave active learning methods for future work.

\subsection{Numerical implementation in finite-sample cases}
We consider the detailed discretization method of the population risk in \eqref{eqn:rp} that requires only training data $(x_i, y_i=f^*(x_i))_{i=1}^N$ sampled from $\rho$.
In general, we apply kernel density estimation methods to approximate the reweighting coefficients in \eqref{eqn:rp} from training data and algorithm support, while the choice of hyperparameters, including kernel variances, will be discussed in the supplementary materials.

\textbf{Approximating the empirical sampling measure.}
The sampling measure $\rho$ can be approximated by an empirical probability measure ${\rho}_{N}$
\begin{equation}\label{eqn:rho}
\frac{\mathrm{d}{\rho}_{N}}{\mathrm{d}x}(x) = \frac{1}{N}\sum_{i=1}^N \kappa(x, x_i).
\end{equation}
Note that \eqref{eqn:rho} is only evaluated on training data,
and $\frac{\mathrm{d}{\rho}_{N}}{\mathrm{d}x}(x_i)$ can be calculated before training.

\textbf{Stratifying the algorithm support.}
With a surrogate model $f_\theta$, we can implement the algorithm $\mathcal{A}$ and determine the algorithm support $\mathcal{S}(f_\theta)$ as $[\tilde{x}_j]_{j=1}^{J}$.
The stratification factor assigned to each support point is calculated as
\begin{equation}\label{eqn:nutheta}
\alpha_\theta(\tilde{x}_j) \propto \left(\frac{\mathrm{d}\nu_\theta}{\mathrm{d}x}(\tilde{x}_j) \right)^{-1},\quad
\frac{\mathrm{d}\nu_\theta}{\mathrm{d}x}(x)
\propto
\dfrac{1}{J}\sum_{j=1}^{J} \kappa(x,\tilde{x}_j).
\end{equation}
where the empirical distribution $\nu_\theta$ of $\mathcal{S}(f_\theta)$ is expressed with a kernel $\kappa$.
This step is necessary if $\frac{\mathrm{d}\nu_\theta}{\mathrm{d}x}$ is far from the uniform distribution.

\begin{remark}
$\nu_\theta$ can be viewed as the task distribution.
While \eqref{eqn:p1} minimizes the mean square error on the sampling measure $\rho$, we are trying to minimize the maximum error on the task distribution $\nu_\theta$.
The discrepancy between \eqref{eqn:p1} and \eqref{eqn:p2} stems from the distribution shift between these two distributions.
In \cref{fig:0}, the red region can be viewed as a visualization of $\nu_\theta$.
\end{remark}

\textbf{Estimating prediction errors on the algorithm support.}
To emphasize support points with large errors, the prediction errors on the algorithm support need to be estimated.
Although we do not have access to $f^*(\tilde{x}_j)$ and therefore $\|f_\theta(\tilde{x}_j)- f^*(\tilde{x}_j)\|_{\mathcal{Y}}^2$,
we can give an estimation based on training data.
The prediction error can be estimated as
\begin{equation}\label{eqn:loss}
\begin{aligned}
&\|f_\theta(\tilde{x}_j)- f^*(\tilde{x}_j)\|_{\mathcal{Y}}^2
\approx \underset{x \sim \kappa(\cdot, \tilde{x}_j)}{\mathbb{E}} \|f_\theta(x)- f^*(x)\|_{\mathcal{Y}}^2  \\
= &\underset{x \sim \rho}{\mathbb{E}} \dfrac{\mathrm{d}\kappa(\cdot, \tilde{x}_j)}{\mathrm{d}\rho}(x) \|f_\theta(x)- f^*(x)\|_{\mathcal{Y}}^2
\approx \dfrac{1}{N} \sum_{i=1}^N \dfrac{\kappa(x_i, \tilde{x}_j)}{\frac{\mathrm{d}{\rho}_{N}}{\mathrm{d}x}(x_i)} \|f_\theta(x_i)- f^*(x_i)\|_{\mathcal{Y}}^2,
\end{aligned}
\end{equation}
which is a reweighted average of training errors.
Therefore, we define an estimated error for $x\in\mathcal{X}$ as
\begin{equation}\label{eqn:ltilde}
l_\theta(x) = \dfrac{1}{N}\sum_{i=1}^N K(x_i, x) \|f_\theta(x_i) - f^*(x_i)\|_{\mathcal{Y}}^2 ,
\quad
K(x_i, x) \propto \dfrac{\kappa(x_i, x)}{\frac{\mathrm{d}{\rho}_{N}}{\mathrm{d}x}(x_i)},
\end{equation}
where $K(x_i, x)$ is a normalized weight
satisfying $\frac{1}{N}\sum_{i=1}^N K(x_i, x) = 1$.

\textbf{Approximating RN derivatives.}
It is important to deal with the reweighting coefficients $\frac{\mathrm{d}\mu_\theta}{\mathrm{d}\rho}$ in \eqref{eqn:rp},
because these RN derivative directly tackle the distribution shift between the input sampling measure and the desired
sample distributions induced by the downstream task.
The coefficients $\omega_\theta(\tilde{x}_j)$ in \eqref{eqn:mut} aim to emphasize points with large prediction errors
in the algorithm support.
Although the accurate errors are unknown, we have calculated the approximate errors $l_\theta(\tilde{x}_j)$.
To approach the supremum in \eqref{eqn:sup}, we can simply define $\omega_\theta(\tilde{x}_j)$ as $\operatorname{softmax}(M \cdot l_\theta(\tilde{x}_j))$.
As $M \geq 0$ goes to infinity, $\mu_\theta$ emphasizes the point with the largest error, so the supremum in \eqref{eqn:sup} can be approached.
However, the task-specific population risk $R$ in \eqref{eqn:rp} will become ill-conditioned for an overly large $M$.
In practical computations, a fixed $M \geq 0$ is taken without much sacrifice of the numerical stability.
On the other hand, for a constant $\omega_\theta(\tilde{x}_j)$, $R$ approximates the mean error on the algorithm support, which is often more robust for optimization.
Generally, we take $\omega_\theta(x)$ as a balance between the mean and supremum error,
\begin{equation}\label{eqn:meansup}
\omega_\theta(\tilde{x}_j) = J\operatorname{softmax}(M \bar{l}_\theta^{-1} \cdot l_\theta(\tilde{x}_j)) + \omega_0.
\end{equation}
where $\omega_0\geq 0$ is a hyperparameter for numerical stability.
The scaling parameter $\bar{l}_\theta$ is the average of $l_\theta(\tilde{x}_j)$ for $\tilde{x}_j \in \mathcal{S}(f_\theta)$.
From the above derivations, we obtain an approximation of the RN derivatives $\frac{\mathrm{d}\mu_\theta}{\mathrm{d}\rho}(x_i)$ as
\begin{equation}\label{eqn:mtheta}
m_\theta(x_i) \propto \dfrac{\mathrm{d}\mu_\theta}{\mathrm{d}\rho_N}(x_i) =\dfrac{1}{J}\sum_{j=1}^{J} {\omega_\theta(\tilde{x}_j)\alpha_\theta(\tilde{x}_j)\kappa(x_i,\tilde{x}_j)}\left(\frac{\mathrm{d}{\rho}_{N}}{\mathrm{d}x}(x_i)\right)^{-1}.
\end{equation}

\textbf{Task-specific empirical risk. }
Consequently, the task-specific empirical risk for training is
\begin{equation}\label{eqn:rn}
R_{N}(\theta) = \dfrac{1}{N}\sum_{i=1}^N m_\theta(x_i) \|f_\theta(x_i)-y_i\|_{\mathcal{Y}}^2 .
\end{equation}
Note that it is necessary to ensure that $m_\theta(x_i)$ is a normalized weight satisfying
$\frac{1}{N}\sum_{i=1}^{N} m_\theta(x_i) = 1$,
because a totally wrong model $f_\theta$ may lead all points in $\mathcal{S}(f_\theta)$ far from all the training data.
As a result, $\kappa(x_i, \tilde{x}_j)$ would be extremely small, so all $\mu_\theta(x_i)$ are close to zero in this finite-sampling case, and the empirical risk $R_N$ could be rather close to zero in absence of this normalization.

\subsection{Training procedure}
The task-specific empirical risk \eqref{eqn:rn} can be minimized using various optimization tools if the gradient of reweighting coefficients $\nabla_\theta m_\theta(x_i)$ is available.
However, this often requires a differentiable algorithm $\mathcal A$ and utilizes back propagation through it, so it could be difficult or expensive to calculate these gradients for general algorithms.
On the other hand, $\nabla_\theta m_\theta(x_i)$ would be ill conditioned since it approximates the supremum error, so the learning rate should be very small to stabilize the training procedure.
Also, the task-specific population risk $R$ could be highly nonlinear, possibly with multiple minima existing.
Therefore, we treat the algorithm $\mathcal{A}$ as a black box for generality and simplicity, and develop another method without querying the gradient of the algorithm implementation.

\textbf{Initializing the model.}
A randomly-initialized model without training generally leads to large errors on the algorithm support,
so it is not useful to improve the prediction accuracy on this support.
The minimization of MSE in \eqref{eqn:p1} is easily achieved,
so an initial model $f_{\theta_{\mathrm{MSE}}} \in \mathcal{F}$ is obtained from \eqref{eqn:p1} to serve as a starting point for subsequent training.
Although this model may not be accurate, it should hopefully give an algorithm support that is not fundamentally incorrect either.

\textbf{Training direction.}
Assume that our training starts from a surrogate model $f_{\theta_0}$.
The current model $f_{\theta_0}$ provides an algorithm support $\mathcal{S}(f_{\theta_0})$, and reweighting coefficients $m_{\theta_0}$ are calculated to characterize which samples should be emphasized in training.
We hope that the surrogate model can predict more accurately on this algorithm support after training, so we can intuitively optimize a reweighted empirical risk $L_N(\theta; m_{\theta_0})$ with respect to $\theta$, while the reweighting coefficients are fixed as $m_{\theta_0}$.
The reweighted empirical risk with fixed reweighting coefficients $m$ is defined as
\begin{equation}\label{eqn:reweighted}
L_N(\theta; m) = \dfrac{1}{N}\sum_{i=1}^N m(x_i) \|f_\theta(x_i)-y_i\|_{\mathcal{Y}}^2.
\end{equation}
In other words, instead of the exact negative gradient direction of $R_N(\theta)$,
we apply an update direction for parameters $\theta$ as
\begin{equation}\label{eqn:direction}
-\frac{1}{N}\sum_{i=1}^N m_{\theta_0} (x_i) \nabla_\theta \|f_\theta(x_i) - y_i\|_{\mathcal{Y}}^2
\end{equation}
for training.
This direction avoids the calculation of gradients of the reweighting coefficients $m_\theta$ and
the training is easy to implement with sample reweighting in many Python packages, for example, Tensorflow  \cite{abadi2016tensorflow}, which is used in this paper.

\textbf{Iterative optimization.}
During training, the reweighting coefficients $m_\theta$ continue to change for two reasons.
On the one hand, the training would significantly change the prediction accuracy of the surrogate model on the previous support $\mathcal{S}(f_{\theta_0})$.
Assume that $\tilde{x}_a$ has the largest error in the previous support $\mathcal{S}(f_{\theta_0})$, which can also be captured by the approximation error $l_\theta$.
By training the model with the reweighted empirical risk $L_N(\theta, m_{\theta_0})$, the prediction error in $\tilde{x}_a$ should be significantly reduced, while the errors in other data points in the previous support can increase as a trade-off.
Therefore, we should not minimize $L_N(\theta, m_{\theta_0})$ excessively towards its minimizer in order to control the errors on other points.
Hopefully, the maximum error on $\mathcal{S}(f_{\theta_0})$ can be reduced after training, and the new surrogate model can improve the prediction accuracy on the previous support.
On the other hand, after training, the algorithm may evaluate the new surrogate model on a different algorithm support than the previous one.
Although the surrogate model can have a more accurate prediction on $\mathcal{S}(f_{\theta_0})$, we cannot guarantee how it performs on the new support.
Based on \cref{ass:1}, if the model changes slightly, the support should also remain basically unchanged.
This inspires us that the reweighting coefficients $m_\theta$ should be updated frequently during training.
This numerical technique is widely used in compressive sensing, dictionary learning, and adversarial training
\cite{yang2011alternating,agarwal2016learning,shaham2018understanding}.

In our training, we apply an iterative optimization method to learn the surrogate model by optimizing the empirical risk $R_N$ gently.
The training procedure consists of updating reweighting coefficients $m_\theta$ based on the current surrogate model and optimizing the reweighted empirical risk $L_N(\theta, m)$ with fixed reweighting coefficients $m=m_\theta$.
It is essential to start training with gentle optimizers, e.g. stochastic gradient descent (SGD) methods with small learning rates and a few training epochs,
while the optimizer should be adjusted according to the optimization performance in the previous iteration.
If the training can reduce the prediction error on the algorithm support, we can adopt optimizers with
more drastic updates.
Otherwise, the optimization should be implemented more cautiously.
This issue will be addressed in the supplementary materials.

\textbf{Early stopping.}
During training, we aim to improve the worst accuracy in the algorithm support \eqref{eqn:sup}.
However, in the fixed-data cases there is no easy way to obtain validation data,
so an alternative early stopping method is applied.
Since we are trying to minimize an empirical risk $R_{N}$, we can monitor $R_{N}$ to apply an early stopping strategy.
Intrinsically, we are trying to minimize $R_{\mathcal{S}}$ in \eqref{eqn:rs}, while $R_N$ serves as an empirical approximation from change of measure.
Therefore, we can define a direct approximation of $R_{\mathcal{S}}$ as
\begin{equation}\label{eqn:rsn}
R_{\mathcal{S},N}(\theta) = \max_{x\in \mathcal{S}(f_\theta)} l_\theta(x) = \max_{j=1,\cdots,J} l_\theta(\tilde{x}_j),
\end{equation}
which is a reliable candidate for early stopping.
After training, the model with the lowest $R_{\mathcal{S},N}$ is accepted as the surrogate model for the downstream task, indicating that it has the lowest maximum prediction error on the corresponding support.

We summarize the overall training algorithm in \cref{alg}, and
the algorithm implementation and numerical examples are available at
\href{https://github.com/MLDS-nus/task-specific-learning}{GitHub}\footnote{\href{https://github.com/MLDS-NUS/task-specific-learning}{https://github.com/MLDS-NUS/task-specific-learning}}.
\begin{algorithm}
\caption{Task-specific supervised training algorithm}\label{alg}
\begin{algorithmic}[0]
\State \textbf{Given} algorithm support $\mathcal{S}$, hypothesis space $\mathcal{F}=\{f_\theta: \mathcal{X}\to\mathcal{Y} \;|\; \theta\in\Theta\}$.
\State Training data $(x_i, y_i)_{i=1}^N$, kernel $\kappa$ with variance $\epsilon^2$, hyperparameters $M$, $\omega_0$.
\Require Initial model parameters $\theta$.
\State Calculate $\rho_N$ as \eqref{eqn:rho} with $\kappa$; initialize Opt as SGD.
\While{stopping criteria not met}
\State $m \gets m_\theta$ as \eqref{eqn:mtheta} with $\mathcal{S}(f_{\theta}), \alpha_\theta, l_\theta,  \omega_\theta, \kappa,M,\omega_0$.\\
\hfill{(Calculate normalized reweighting coefficients from surrogate model $f_\theta$)}
\State Update Opt; $\theta \gets$ Opt $L_N(\theta;m) = \frac{1}{N}\sum_{i=1}^N m(x_i) \|f_\theta(x_i)-y_i\|_{\mathcal{Y}}^2$ as \eqref{eqn:reweighted}.\\
\hfill{(Update model by training the reweighted empirical risk with Opt)}
\EndWhile
\Ensure $\theta$ with the minimal $R_{\mathcal{S},N}(\theta)$.
\end{algorithmic}
\end{algorithm}

\section{Numerical experiments}
In our experiments, we compare the results of MSE minimization \eqref{eqn:p1}
and the task-specific (TS) supervised learning problem \eqref{eqn:rp},
where the results of model parameters are denoted by $\theta_{\text{MSE}}$ and $\theta_{\text{TS}}$, respectively.
In each case, the training dataset size is $N=10^5$ and 20 random seed are tested.
We demonstrate the following conclusions from our numerical results.
\begin{enumerate}
\item The surrogate model $f_{\theta_{\text{TS}}}$ have a better accuracy on its algorithm support, i.e. $R_{\mathcal{S}}(\theta_{\text{TS}}) < R_{\mathcal{S}}(\theta_{\text{MSE}})$.
Consequently, we expect that $\theta_{\text{TS}}$ gives a more accurate algorithm output $\mathcal{A}(f_{\theta_{\text{TS}}})$, i.e. $J_{\mathcal{A}}(\theta_{\text{TS}}) < J_{\mathcal{A}}(\theta_{\text{MSE}})$.
\item For a small hypothesis space with weak expression ability, the improvement of $J_{\mathcal{A}}(\theta_{\text{TS}})$ compared to $J_{\mathcal{A}}(\theta_{\text{MSE}})$ is significant.
As the hypothesis space becomes larger with stronger expression ability, this improvement would weaken.
\item The performance of the task-specific supervised learning problem is affected by the distribution mismatch between
the sampling environment $(\rho)$ and the task environment ($\mathcal{S}(f^*)$).
If the distribution shift is large, the improvement between $J_{\mathcal{A}}(\theta_{\text{TS}})$ and $J_{\mathcal{A}}(\theta_{\text{MSE}})$ is significant.
Our method would not be very useful if the sampling measure $\rho$ is close to the task distribution of $f^*$.
\end{enumerate}

\subsection{Multistep prediction of dynamical systems}\label{nm1}
In this example, we consider using data of one-step flow to predict a multistep
flow trajectory from a given initial point of an autonomous dynamical system.

\textbf{Downstream task. }
Consider the Lorenz system \cite{lorenz1963deterministic},
\begin{equation}\label{eqn:ex1ode}
\left\{
\begin{array}{l}
\dot{x}_1  = C_\sigma (x_2 - x_1) ,\\
\dot{x}_2  = x_1 (C_\rho - x_3) - x_2 ,\\
\dot{x}_3  = x_1x_2 - C_\beta x_3,
\end{array}
\right.
\end{equation}
with parameters $C_\rho=28$, $C_\sigma=10$, and $C_\beta=8/3$,
and we have access to the initial and final states of some short trajectories over
a time interval of length $\tau=0.01$.
Our downstream task is to predict a multistep trajectory $[f^*(x_0), f^{*2}(x_0), \cdots, f^{*N_T}(x_0)]$
over a larger
time interval $[0,N_T\tau]$ consisting of ${N_T}$ time steps (each of length $\tau$)
from a given initial condition $x_0$.
Here, the ground truth $f^*(x)$ is the flow map from an initial point
$x\in\mathcal{X}$ over a time step $\tau$,
and $f^{*n}(x_0)$ represents the composition function as $f^*\circ f^*\circ \cdots \circ f^*(x)$
where the function $f^*$ is applied $n$ times.

\textbf{Downstream task algorithm. }
With a surrogate model $f$ of the flow map $f^*$, the predicted multistep trajectory would be simply calculated as $\mathcal{A}(f) = [f(x_0)$, $f^{2}(x_0)$, $\cdots, f^{N_T}(x_0)]$.
Therefore, the algorithm support would be $\mathcal{S}(f) = [x_0$, $f(x_0)$, $f^{2}(x_0)$, $\cdots$, $f^{N_T-1}(x_0)]$ $(J=N_T)$.
Our aim is to minimize the average discrepancy along the multistep trajectory,
\begin{equation}
J_{\mathcal{A}}(\theta) = \|\mathcal{A}(f_\theta) - \mathcal{A}(f^*)\| =
\dfrac{1}{J} \sum_{j=1}^J \|f_\theta^{j}(x_0) - f^{*j}(x_0)\|_2.
\end{equation}

\textbf{Training data generation. }
The sampling measure $\rho$ consists of two parts $\rho = (1-\alpha)\rho_0 + \alpha \rho_1$.
$\rho_0$ is the uniform measure $\mathcal{U}([-25,25]^2\times [0,50])$, and $\rho_1$ is an empirical task distribution
\begin{equation}\label{eqn:1rho1}
\rho_1 = \frac{1}{N_T} \sum_{x\in \mathcal{S}(f^*)} \kappa(\cdot,x),
\end{equation}
which presents training data near the correct algorithm support.
The parameter $\alpha$ will be varied to enable an ablation study of the distribution shift.
We simulate \eqref{eqn:ex1ode} using robust ODE integrators \cite{hairer1996solving}
with small step sizes to obtain labels with high precision.
The kernel variances in \eqref{eqn:rho}, \eqref{eqn:nutheta},
\eqref{eqn:ltilde}, \eqref{eqn:mtheta}, and \eqref{eqn:1rho1} are taken as 1.0, 2.0, 1.0, 1.0, 1.0, respectively.

\textbf{Hypothesis space.}
We apply a ResNet \cite{he2016deep} with one hidden layer of $d$ neurons,
$f_\theta(x) = x + \tau (w_1\sigma(w_0 x + b_0) + b_1)$,
to approximate the flow map $f^*$,
where the parameters $\theta=(w_0, b_0, w_1, b_1)$ are
$w_0\in \mathbb{R}^{d\times 3}, w_1\in \mathbb{R}^{3\times d}, b_0 \in \mathbb{R}^d, b_1\in \mathbb{R}^3$.
The activation function $\sigma$ in all three examples is the exponential linear unit (ELU) \cite{clevert2015fast}.

\begin{figure}[htb]
    \centering
    \includegraphics[width=\linewidth]{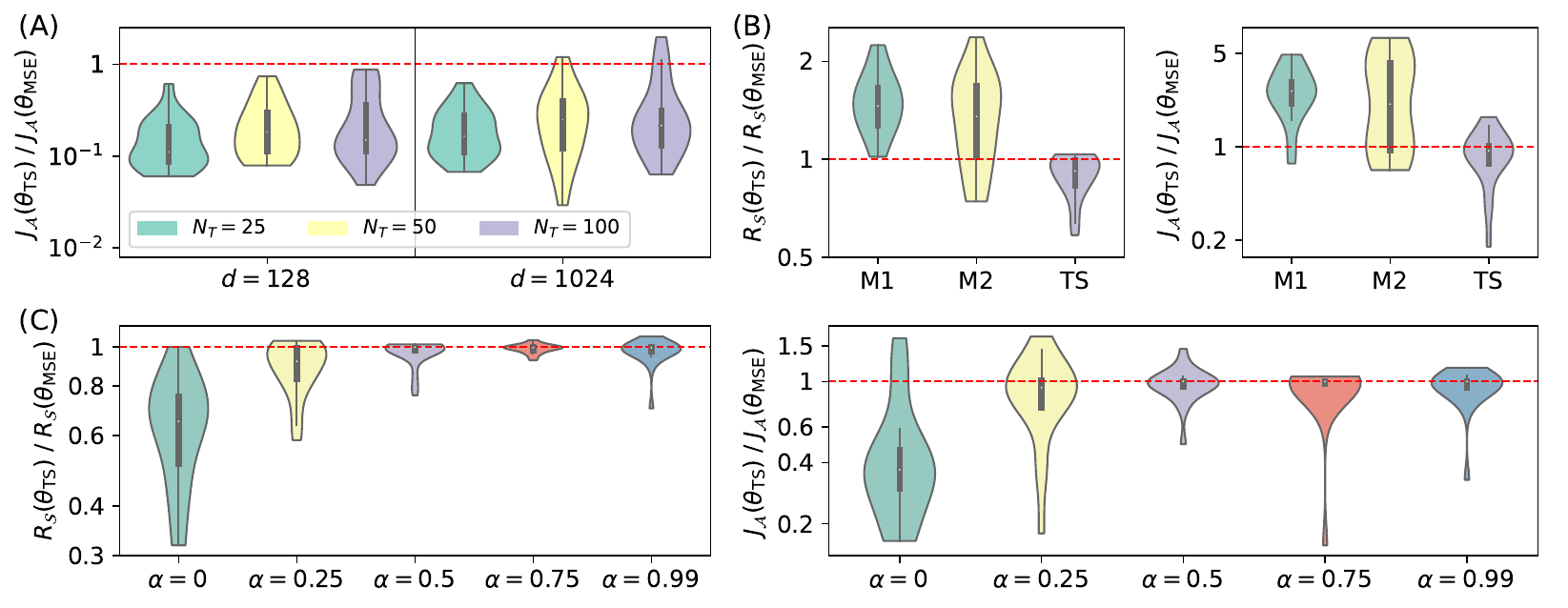}
    \caption{Results of the multistep prediction of dynamical systems.
    The horizontal red dashed line indicates the absence of any improvement,
    below which improvements are achieved.
    In each case, the violin plot of 20 runs with different random seeds are shown.
    (A) The improvement of the algorithm output error $J_{\mathcal{A}}$ for different $d$ and $N_T$.
    (B) Ablation study of different reweighting methods.
    The improvements of the maximum prediction error on the algorithm support $R_{\mathcal{S}}$
    and the algorithm output error $J_{\mathcal{A}}$ are shown.
    (C) Ablation study of different distribution shift $\alpha$.}
    \label{fig:1}
\end{figure}

\textbf{Results and discussion.}
In our numerical experiments, we randomly generate 20 different downstream tasks including training data ($\alpha=0$) and the initial condition $x_0$ (sampled from the uniform measure $\mathcal{U}([-12.5,12.5]^2\times[12.5,37.5])$).
The hidden layer of neural networks in the hypothesis space has $d=128$ or $1024$ neurons, and the prediction time step is taken as $N_T=25,50$ or $100$.
Then, the improvements of our method is illustrated by $J_{\mathcal{A}}(\theta_{\text{TS}}) / J_{\mathcal{A}}(\theta_{\text{MSE}})$ in \cref{fig:1}(A).
By minimizing MSE, $\theta_{\text{MSE}}$
would be the optimal one for the single-step prediction error over the sampling measure,
but it is not generally the optimal one for multistep prediction starting from a specific initial condition.
By minimizing the task-specific empirical risk $R_N$,
the surrogate model $f_{\theta_{\text{TS}}}$ predicts well around the true trajectory,
so our method can significantly reduce the algorithm output error $J_{\mathcal{A}}$ in most cases.
As the time period $N_T$ increases, the output error of the MSE model $J_{\mathcal{A}}(\theta_{\text{MSE}})$
increases due to error accumulation, while $\theta_{\text{TS}}$ still generally outperforms $\theta_{\text{MSE}}$.

\textbf{Ablation study: Reweighting methods.}
To solve the task-specific supervised learning problem \eqref{eqn:rp}, we train the model by minimizing a reweighted MSE.
The reweighted coefficients are determined according to the downstream task and continue to change during model updating.
As an ablation study, we show that this reweighting strategy is essential for the improvement of the algorithm output,
and classical downstream task-agnostic reweighting schemes are not effective.
In particular, we choose the following natural reweighting methods:\\
\begin{enumerate}[label=M\arabic*:, leftmargin=1.0cm]
    \item Minimize an MSE on the uniform measure, so the reweighting coefficients are $m(x_i) = (\frac{\mathrm d\rho_N}{\mathrm dx}(x_i))^{-1}$.
    \item Fine-tune the surrogate model by attaching larger weights to samples with larger error, so the reweighting coefficients are $m(x_i)=\|f_\theta(x_i) - y_i\|_{\mathcal{Y}}^2$.
\end{enumerate}
We consider the $\alpha=0.25$ case and the results are shown in \cref{fig:1}(B).
We observe that since these methods do not take the downstream task into consideration,
they cannot achieve a better performance in the algorithm output.

\textbf{Ablation study: distribution shift. }
In this ablation study,
we consider a particular case of $N_T=100$ and $d=128$,
and the sampling measure is varied as $\alpha=0, 0.25, 0.5, 0.75, 0.99$ to obtain different distribution shift.
For $\alpha=0.99$, most of the training data is concentrated around $\mathcal{S}(f^*)$,
which is useful for the task, so the MSE model $f_{\theta_{\text{MSE}}}$ is expected to give
a rather accurate prediction on $\mathcal{S}(f^*)$.
Therefore, the output error $J_{\mathcal{A}}(\theta_{\text{MSE}})$ should be small.
Under this setting, our method cannot give a significantly better surrogate model for the downstream task.
As $\alpha$ decreases, the distribution shift between the sampling measure $\rho$ and the correct task distribution becomes larger,
so the prediction error of $f_{\theta_{\text{MSE}}}$ on the algorithm support $\mathcal{S}(f_{\theta_{\text{MSE}}})$ increases.
With the increase of $R_{\mathcal{S}}(\theta_{\text{MSE}})$ and $J_{\mathcal{A}}(\theta_{\text{MSE}})$,
the improvement of $f_{\theta_{\text{TS}}}$ is more significant,
which can be observed from both the maximum error on the support
$R_{\mathcal{S}}$ and the algorithm output error $J_{\mathcal{A}}$ in \cref{fig:1}(C).
This confirms that our method is more effective when the distribution shift is larger.

\subsection{Tracking problem}
\label{nm2}
The next example goes well beyond the previous where the downstream task is
an inverse problem involving optimal control.
Concretely, we consider learning the dynamics of a control system
with the goal of solving a downstream tracking problem.

\textbf{Downstream task. }
Consider the pendulum on a cart as a standard example in optimal control \cite{grune2017nonlinear}, governed by a dynamical system,
\begin{equation}\label{eqn:ex2ode}
\left\{
\begin{array}{l}
\dot{x}_1 = x_2,\\
\dot{x}_2 = f(x_1, x_2, u),\\
\dot{x}_3 = x_4,\\
\dot{x}_4 = u.\\
\end{array}
\right.
\end{equation}
In \eqref{eqn:ex2ode}, $x_1(t), x_2(t)\in \mathbb{R}$ are the angle and the angular velocity of the pendulum, while $x_3(t), x_4(t)\in \mathbb{R}$ are the position and the velocity of the cart, respectively.
Assume that the system is known to be governed by \eqref{eqn:ex2ode} without the knowledge of the exact expression of the angular acceleration $f$.
The ground truth of $f$ in \eqref{eqn:ex2ode} is
\begin{equation}\label{eqn:ex2fs}
f^*(x_1, x_2, u)= -\frac{g}{l} \sin x_1 - u\cos x_1 - \frac{k_L}{l} x_2 |x_2| - k_R \operatorname{sgn}x_2,
\end{equation}
with fixed parameters $g=9.81$, $l=10$, and $k_R=k_L=0.01$.
Here, we only have access to samples of $f^*$ at some discrete points in the state-action space.
Our downstream task is to solve a tracking problem over $t\in[0,T=1]$:
\begin{equation}\label{eqn:ex2cu}
\min_{u\in \mathcal{U}} \quad C(u) = \int_0^T \|x(u;f^*)(t) - x_r(t)\|^2 \mathrm{d}t.
\end{equation}
The reference trajectory for tracking is set as $x_r(t) = (\pi+1-t,t-1,0,0), t\in[0,T]$,
representing that the pendulum moves from $\pi+1$ to the upright position $\pi$ with velocity changes from $-1$ to $0$ linearly, while the position and the velocity of the cart remains zero.
Here, $x(u;f)$ denotes the solution of \eqref{eqn:ex2ode} with control input $u(t)$ from the initial condition $x(t=0)=x_r(0)$.
The control space $\mathcal{U}$ is a set of some piecewise constant functions in $t\in[0,T]$.
Concretely, for $u\in\mathcal{U}$, we have $u(t)$ is a constant in $[-10,10]$ for $t\in [(i-1)\tau, i\tau)$ for $i=1,\cdots,N_T$ with $N_T=20$ and $\tau=T/N_T$.

\textbf{Downstream task algorithm.}
Denote $\hat{x}(u;f)(i\hat{\tau})$ as the numerical solution of \eqref{eqn:ex2ode} obtained from an explicit Euler scheme with a refined time step size $\hat{\tau}=\tau/N_\tau$ with $N_\tau=10$.
The downstream task algorithm $\mathcal{A}$ for the tracking problem \eqref{eqn:ex2cu} gives the optimal control $u_\theta$ by directly minimizing
a discretized version of \eqref{eqn:ex2cu} with a surrogate model
$f_\theta \in \mathcal{F}$ as the unknown angular acceleration $f$:
\begin{equation}\label{eqn:ex2ut}
u_\theta := \argmin_{u\in\mathcal{U}} \; C_\tau(u; f_\theta) = \sum_{j=1}^{N_T} \tau\|\hat{x}(u;f_\theta)(j{\tau}) - x_r(j{\tau})\|^2,
\end{equation}
which can be solved with various optimization methods.
With the ground truth, the accurate optimal control can be obtained
as $u^* := \argmin_{u\in\mathcal{U}} \; C_\tau(u; f^*)$.
We assume that based on a surrogate model $f_\theta$,
the control input $u_\theta$ is obtained by solving \eqref{eqn:ex2ut}.
The algorithm support $\mathcal{S}(f_\theta)$ is defined as where the surrogate model
is evaluated during the explicit Euler scheme with the control input $u_\theta$ ($J=N_T N_\tau$):
\begin{equation}\label{eqn:ex2sf}
\mathcal{S}(f_\theta)=[\left(\hat{x}_1(u_\theta;f_\theta)(j\hat{\tau}), \hat{x}_2(u_\theta;f_\theta)(j\hat{\tau}), u_\theta(j\hat{\tau})\right)]_{j=0}^{N_T N_\tau - 1}.
\end{equation}
Since we are only concerned with the optimization result,
the states visited by the optimization process need not be included in the algorithm support.
Nevertheless, the surrogate model should be close to the ground truth in the
neighbourhood of the support, so the optimization process should be locally stable near the result.
We can apply derivative-free methods to solve the optimization problem \cref{eqn:ex2ut}, which accords with \cref{ass:1}.
On the other hand, we can also apply optimization methods that require derivatives of the surrogate model, because the algorithm output only depends on the surrogate function evaluations on the support and is independent of the optimization process.
The final algorithm output for comparison is the tracking error of the ground truth
\begin{equation}\label{eqn:ex2sftt}
\mathcal{A}(f_\theta) = C_\tau(u_\theta; f^*) \geq C_\tau(u^*; f^*) = \mathcal{A}(f^*),
\end{equation}
and we have the algorithm output error as $J_{\mathcal{A}}(\theta) = \mathcal{A}(f_\theta) - \mathcal{A}(f^*)$.

\textbf{Training data generation. }
The input of the training data is sampled from the uniform measure
$(x_1, x_2, u)\sim \mathcal{U}([0,2\pi]\times[-5,5]\times[-11,11])$
and then $u$ is clipped to the interval $[-10,10]$ to obtain additional control boundary data.
We have accurate acceleration $f^*(x_1, x_2, u) \in \mathcal{Y} = \mathbb{R}$ as labels of training data.
The kernel variances in \eqref{eqn:rho}, \eqref{eqn:nutheta}, \eqref{eqn:ltilde},
and \eqref{eqn:mtheta} are taken as 0.01, 0.01, 0.01, 0.05, respectively.

\textbf{Hypothesis space. }
The hypothesis space is either multivariate polynomials of $x_1, x_2, u$ with degree at most $d$ (Poly$d$), or fully-connected neural networks with a single hidden layer of $d$ neurons (FNN$d$).
We consider hypothesis spaces of different expressive abilities
in order to illustrate the impact of hypothesis space sizes.

\begin{figure}[htb]
    \centering
    \includegraphics[width=\linewidth]{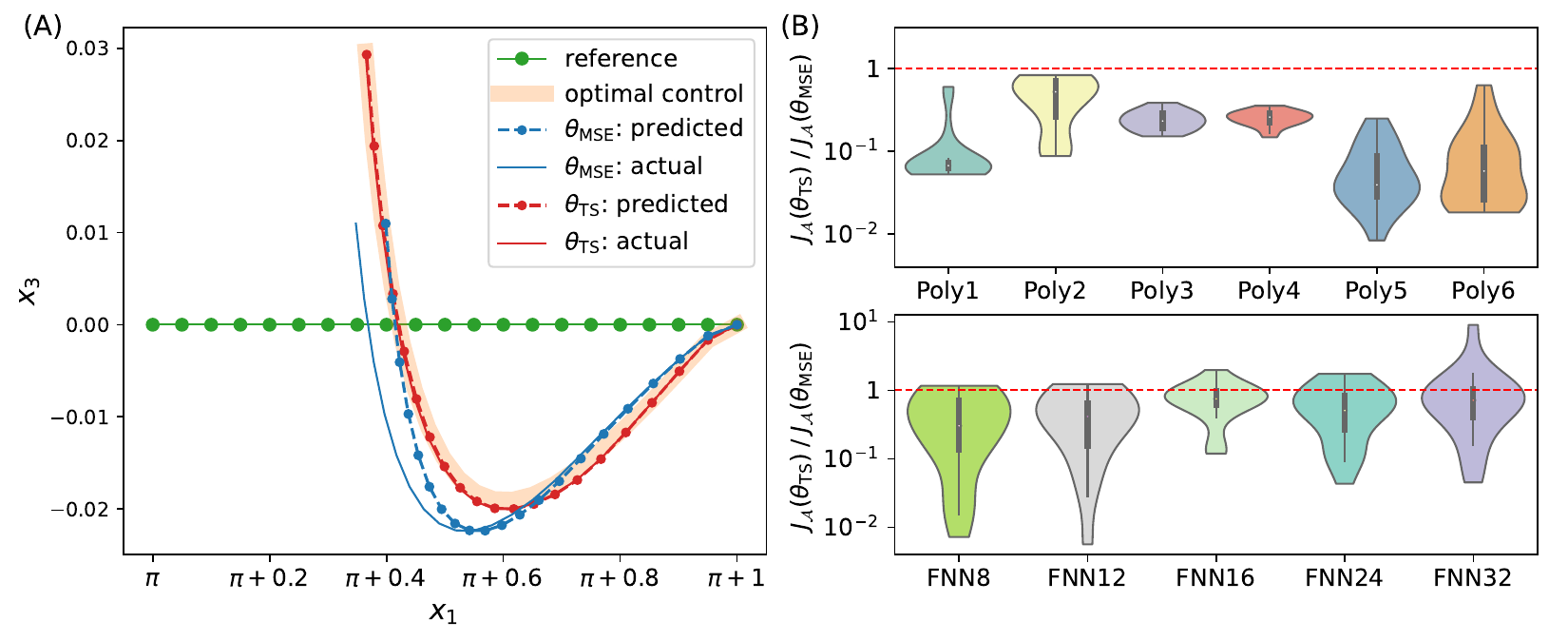}
    \caption{Results of the tracking problem.
    (A) As a typical result, for $\theta=\theta_{\mathrm{MSE}}$ (blue) and $\theta_{\mathrm{TS}}$ (red),
    the predicted trajectory $x(u_\theta, f_\theta)$ (dashed lines with dots) and the actual trajectory
    $x(u_\theta, f^*)$ (solid lines) of the corresponding optimal control input $u_\theta$
    are shown in the $(x_1, x_3)$ plane, respectively.
    (B) Ablation study of different hypothesis space sizes.
    The improvements of $J_{\mathcal{A}}$ for Poly$d$ (upper)
    and FNN$d$ (lower) are shown. }
    \label{fig:2}
\end{figure}

\textbf{Results and discussion. }
The surrogate model $f_\theta$ is only evaluated on the trajectory of the
optimal control $\mathcal{S}(f_\theta)$, so it should be accurate near this
trajectory.  In order to obtain a better control input, the trajectory predicted
by the surrogate model $x(u_\theta;f_\theta)$ should be close to the actual
trajectory $x(u_\theta;f^*)$.  However, this trajectory is related by the
undetermined control input $u_\theta$, so this problem is more complicated than
the previous one.

Here, we show a typical example of numerical results in \cref{fig:2}(A),
with neural networks of width $d=8$ as the hypothesis space.
For $\theta=\theta_{\mathrm{MSE}}$ and its optimal control $u_\theta$,
there exists a visible deviation between the predicted trajectory $x(u_\theta; f_\theta)$ and the actual trajectory $x(u_\theta; f^*)$, because the approximation error between $f_\theta$ and $f^*$ is relatively large in the neighbourhood of $\mathcal{S}(f_\theta)$.
For $\theta=\theta_{\mathrm{TS}}$, the predicted trajectory $x(u_\theta; f_\theta)$ of this optimal control $u_\theta$ looks identical to $x(u_\theta; f^*)$, and therefore the optimal control solved from $\theta=\theta_{\mathrm{TS}}$ is more accurate.
Consequently, the actual trajectory of $\theta_{\mathrm{TS}}$ is similar to the optimal control trajectory $x(u^*; f^*)$.

\textbf{Ablation study: hypothesis space. }
To demonstrate the impact of the hypothesis space sizes, we consider different width $d$ in the neural networks, and for each case, we apply 20 different random seeds to generate training data and initialize the parameters of the neural networks.
Our improvements are shown by $J_{\mathcal{A}}(\theta_{\text{TS}}) / J_{\mathcal{A}}(\theta_{\text{MSE}})$ in \cref{fig:2}(B).
For Poly$d$, as the degree of polynomials $d$ increases, the hypothesis space has a stronger expressive ability, and both $J_{\mathcal{A}}(\theta_{\text{MSE}})$ and $J_{\mathcal{A}}(\theta_{\text{TS}})$ decrease, while our method presents a more accurate output in each case.
For FNN$d$ with $d=8, 12, 16, 24, 32$, with more constrained model sizes, the improvement of our method is more significant.
Note that for each Poly$d$, $\theta_{\text{MSE}}$ is almost the same for different random seeds because $\cref{eqn:p1}$ is a convex problem.
In contrast, different random initializations of neural networks give totally different model parameters $\theta_{\text{MSE}}$ and the output error $J_{\mathcal{A}}$.
Therefore, we can see the impact of hypothesis spaces in the lower panel of \cref{fig:2}(B).

\subsection{MEP calculation}
\label{nm3}
Our final example concerns a well-known problem in computational chemistry and material science.
Given a potential energy function $E$ with multiple local minima,
an important task is to calculate the minimum energy path (MEP) \cite{weinan2002string} connecting two metastable states.
These paths are crucial for understanding the transition between different states of a system,
including the identification of transition states and the calculation of energy barriers \cite{weinan2010transition}.

In modern data-driven computational chemistry, the potential energy surface $E$
(and its gradient $\nabla E$) is often approximated
by machine learning models \cite{chmiela2017machine, zhang2018deep, doerr2021torchmd, unke2021machine, zhang2024pretraining}, which are trained on data generated from density functional theory calculations.
In this case, an MSE-type loss function is typically used to train the model.
However, the purpose of learning this energy function is not to minimize the MSE of the energy or force prediction
on the training sampling measure,
but rather to use it for downstream tasks such as the calculation of the MEP.
Consequently, from our formulation it is clear
that the MSE loss function may not be the most appropriate choice for training the model.
Here, we demonstrate using a minimal example how our method can be used to
improve the accuracy and robustness of this approach.

\begin{figure}[htbp]
    \centering
    \includegraphics[width=\linewidth]{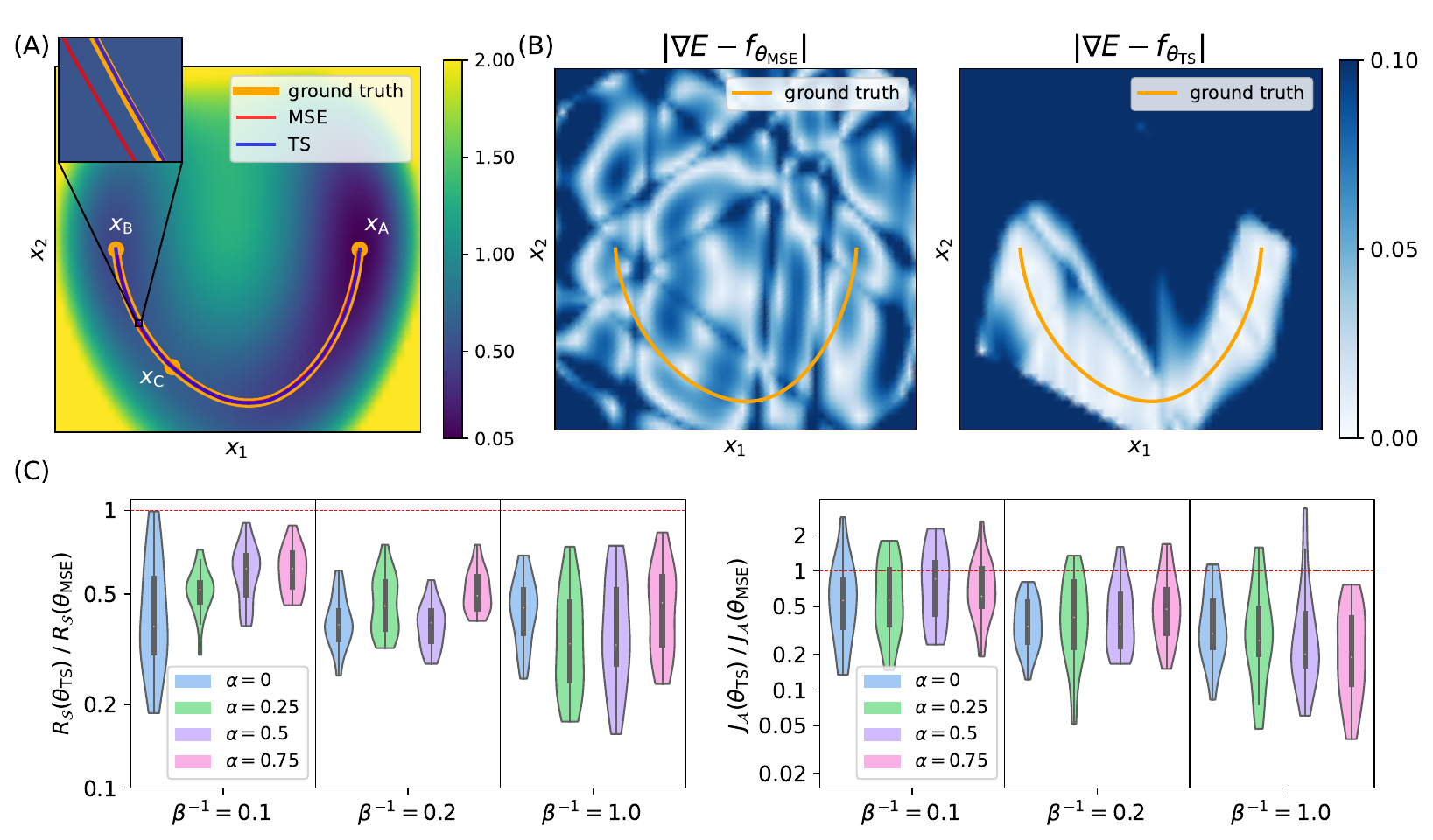}
    \caption{
    Results of the MEP calculation problem.
    (A) The energy landscape $E$ on $x_3=0$, with the correct MEP (orange), and a typical result of the MEP obtained by MSE (red) and TS (blue).
    (B) The prediction errors of $f_{\theta_{\mathrm{MSE}}}$ and $f_{\theta_{\mathrm{TS}}}$ in this typical result on $x_3=0$.
    (C) Ablation study of different distribution shift $\alpha$.
    The improvements of $R_{\mathcal{S}}$ and $J_{\mathcal{A}}$ are shown. }
    \label{fig:3}
\end{figure}

\textbf{Downstream task. }
Consider an energy function $E$ on $\mathbb{R}^3$ with the form of
\begin{equation}\label{eqn:ex3ene}
E(x)= (x_1^2-1)^2 + \frac14(x_2^2-1)^2 +\frac12 x_3^2 + x_1^2x_2^2 + \frac{1}{10}(x_1^3-3x_1+x_2^3+3x_2(1-x_1^2)).
\end{equation}
The energy has two minima at $x_{\mathrm{A}}=(1,0,0)$ and $x_{\mathrm{B}}=(-1,0,0)$,
with energy $E(x_{\mathrm{A}}) = 0.05$ and $E(x_{\mathrm{B}})=0.45$,
as shown in \cref{fig:3}(A).
We aim to calculate the MEP connecting two metastable states, which is a smooth path $\phi(\alpha):[0,1]\to \mathcal{X}$ connecting $\phi(0)=x_{\mathrm{A}}$ and $\phi(1)=x_{\mathrm{B}}$.
At each point on the MEP, the tangent direction $\phi'(\alpha)$ is parallel to the gradient direction $\nabla E(\phi(\alpha))$ for $\forall \alpha \in (0,1)$ \cite{weinan2002string}.
The state with the highest energy along the MEP is a critical point where the gradient vanishes, called the transition state.
The energy barrier associated with the MEP characterizes how hard this transition could be.
For this energy function, the MEP between $x_{\mathrm{A}}$ and $x_{\mathrm{B}}$ is associated with a transition state $x_{\mathrm{C}}\approx (-0.5367, -0.9669,0)$, with $E(x_{\mathrm{C}})\approx 0.63$.

\textbf{Downstream task algorithm. }
The string method \cite{weinan2002string,weinan2007simplified} is a well-known algorithm to calculate an MEP between two minima of the energy function, which only requires access to the negative gradient field $f^*=-\nabla E$.
The string method applies a chain of nodes in $\mathcal{X}$ in the configuration space to approximate the MEP.
Given an initial string with $(K+1)$ nodes $\phi_0^{(0)}, \cdots, \phi_K^{(0)}$,
the algorithm of the string method is as follows,
with a surrogate negative gradient field $f_\theta$ to iterate $\phi_k^{(t)}:= \phi_k^{(t)}(f_\theta)$.
\begin{enumerate}
\item String evolution: For $k=0,\cdots,K$, calculate $\hat{\phi}_k^{(t+1)} = \phi_k^{(t)} + \tau f_\theta(\phi_k^{(t)})$;
\item Reparameterize the nodes $[\hat{\phi}_k^{(t+1)}]$ with equal arc lengths to obtain $[{\phi}_k^{(t+1)}]$;
\item Check convergence, otherwise go to Step 1.
\end{enumerate}
Because the downstream task algorithm is iterative based on the current string, we define the algorithm support as the final convergent string:
$\mathcal{A}(f_\theta) = \mathcal{S}(f_\theta) = \lim\limits_{t\to \infty}[\phi_0^{(t)}(f_\theta), \cdots, \phi_K^{(t)}(f_\theta)]$.
Our task is to minimize the discrepancy of the MEP:
\begin{equation}\label{eqn:ex3ja}
J_{\mathcal{A}}(\theta) = \|\mathcal{A}(f_\theta) - \mathcal{A}(f^*)\|
= \dfrac{1}{K+1} \sum_{k=0}^K \lim\limits_{t\to \infty}\|\phi_k^{(t)}(f_\theta) - \phi_k^*\|_2,
\end{equation}
where $\phi_k^*$ represents the correct MEP node.

\textbf{Training data generation. }
As in the case of the Lorenz system, the sampling measure $\rho$ consists
of two parts $\rho = (1-\alpha)\rho_0 + \alpha\rho_1$.
$\rho_0$ is the Boltzmann distribution at a temperature $\beta^{-1}$,
sampled from the overdamped Langevin equation \cite{weinan2010transition},
\begin{equation}\label{eqn:ex3lan}
\mathrm{d}x = -\nabla E(x) \mathrm{d}t + \sqrt{2\beta^{-1}} \mathrm{d}W_t,
\end{equation}
which will relax to the Boltzmann distribution $\rho_0\propto \exp(-\beta E)$ as time progresses.
If the temperature $\beta^{-1}$ is much lower than the energy barrier, most of the data sampled from $\rho_0$ will be concentrated near the metastable states, whereas data near the transition states are relatively rare.
The other part $\rho_1$ is concentrated around the correct MEP nodes,
\begin{equation}\label{eqn:meprho1}
\rho_1 = \dfrac{1}{K+1} \sum_{k=0}^K\kappa(\cdot, \phi_k^*).
\end{equation}
The corresponding label of the training data $x_i$ is calculated as the accurate negative gradient $y_i=f^*(x_i)=-\nabla E(x_i)$ in $\mathcal{Y}=\mathbb{R}^3$.
The kernel variances in \eqref{eqn:rho}, \eqref{eqn:nutheta}, \eqref{eqn:ltilde}, \eqref{eqn:mtheta}, and \eqref{eqn:meprho1} are taken as 0.01, 0.001, 0.001, 0.01, 0.01, respectively.

\textbf{Hypothesis space. }
We construct a parameterized energy function as \cite{yu2021onsagernet},
\begin{equation}\label{eqn:ex3hs}
E_\theta(x) = 10^{-4}\|x\|_2^2 + (w_0^\top x + b_0)^2 + \operatorname{NN}(x), \quad \theta\in\Theta,
\end{equation}
where $\operatorname{NN}$ is a two-layer neural network with $d=32$ hidden neurons
$\operatorname{NN}(x) = w_2^\top\sigma(w_1 x+b_1) + b_2$.
The parameters $\theta=(w_0, b_0, w_1, b_1, w_2, b_2)\in\Theta$ are $w_0 \in\mathbb{R}^3, b_0\in\mathbb{R}, w_1 \in\mathbb{R}^{d\times3}, b_1\in\mathbb{R}^d, w_2 \in \mathbb{R}^{d}, b_2\in\mathbb{R}$.
The first two terms in \eqref{eqn:ex3hs} aim to establish the coercivity of the parametrized energy function.
The ELU activation function ensures that $E_\theta$ is continuously differentiable, and we define the hypothesis space for the surrogate model as $\mathcal{F}=\{-\nabla_x E_\theta(x)| \theta\in\Theta\}$.

\textbf{Results and discussion. }
Here we explain our idea in the situation of rare events.
Because the training data sampled from $\rho_0$ are mainly located around the global minimum point $x_{\mathrm{A}}$, if we treat all samples equally by minimizing MSE, the surrogate model would learn the gradient well around $x_{\mathrm{A}}$.
In contrast, the input points near the transition states will have large prediction errors due to their rarity in the training dataset.
Moreover, with the same sampling measure, this situation will not change with increasing training data size.
However, the data near the transition states are very important to locate the MEP and obtain accurate information on the transition.
Even if some new training data are added near the transition tube, for example, data sampled from $\rho_1$, as long as the data near minimum points are dominant in the training dataset, minimizing MSE could still be a bad choice.
In contrast, with our methodology the training data near the current MEP calculated from string methods will be attached to large reweighting coefficients, so the gradient field around the transition tube would also be learned well.

We generate $10^6$ training data from $\rho_0$ with $\beta^{-1}=0.1$, $0.2$, or $1.0$ and $\rho_1$, and then randomly select training data with $\alpha=0,0.25,0.5,0.75$.
A typical result with $\beta^{-1}=1.0$ and $\alpha=0.25$ is illustrated in \cref{fig:3}(A).
The zoomed-in figure shows that the TS result is more accurate than the MSE result.
\Cref{fig:3}(B) shows that $f_{\theta_{\text{TS}}}$ has a lower prediction error around the MEP as expected.

\textbf{Ablation study: distribution shift.}
For each $\beta^{-1}$ and $\alpha$, we generate different training data and parameter initialization of neural networks.
The improvements of the maximum prediction error on the algorithm support $R_{\mathcal{S}}$ and the algorithm output error $J_{\mathcal{A}}$ are shown in \cref{fig:3}(C).
The surrogate model obtained from task-specific supervised learning problem can obtain a smaller maximum prediction error on the algorithm support in most cases, and consequently, a more accurate MEP.
More importantly, as $\alpha$ decreases, the distribution shift between the training data and the algorithm support becomes larger, so the improvement of our method becomes more significant.
In this example, the relationship between $R_{\mathcal{S}}$ and $J_{\mathcal{A}}$ is rather complicated, so the algorithm output error may perform a little differently.

\section{Conclusion and discussion}
In this paper, we consider learning a predictive model for a given downstream task based on training data sampled from a sampling measure.
We assume that a particular determined downstream task algorithm has been developed based on scientific computing knowledge,
and our goal is to ensure that this algorithm can present an accurate output by using the surrogate model.
First, we show that this problem cannot be generally solved by the classical supervised learning problem which minimizes the mean square error, because of the distribution shift between training and task environments.
This point is often not discussed in existing approaches in machine-learning-augmented scientific computing.

Based on Lipschitz continuity,
we establish an upper bound of the algorithm output error, which represents the maximum prediction error on the algorithm support.
Then, we formulate a task-specific supervised learning problem based on the sampling measure of training data to approximate the maximum prediction error, where Radon--Nikodym derivatives related to the surrogate model are attached as reweighting coefficients.
In the numerical implementation, we apply kernel density estimation to calculate the reweighting coefficients.
To learn the surrogate model, we develop an iterative optimization algorithm with an early stopping strategy.
We present three numerical examples to demonstrate the effectiveness of the task-specific supervised learning problem compared to MSE.
We also conduct ablation studies of distribution shift and hypothesis space sizes.

There are a number of interesting directions for future work.
First, we considered the task-specific supervised learning problem in the expectation sense,
which is in principle applicable to noisy training data and stochastic downstream task algorithms.
It would be of interest to generalize our method to such scenarios.
Second, we currently assume that the training data is fixed and no new data can be sampled and labelled.
In some applications involving coupling of learning and data generation,
it is possible to sample new data \cite{pestourie2020active,lye2021iterative,bash2021.MultiFidelityHighThroughputOptimization,mekki-berrada2021.TwostepMachineLearning}
and update the surrogate model iteratively.
In this case, our approach can be extended to provide a task specific active learning strategy,
which can be used to guide the sampling of new data to improve the surrogate model
for downstream tasks.

\textbf{Acknowledgement.}
This research is supported by the National Research Foundation, Singapore, under the NRF fellowship (project No. NRF-NRFF13-2021-0005).

\bibliographystyle{siam}
\bibliography{references}

\end{document}


\maketitle

\section*{Details in numerical implementations}
In this part, we provide additional details in numerical implementations of our method.

\textbf{Kernel variances.}
To determine the reweighting coefficients $m_\theta$ in \eqref{eqn:mtheta}, we apply kernel density estimation to approximate measures from discretized points in \eqref{eqn:rho}, \eqref{eqn:nutheta}, \eqref{eqn:ltilde}, and \eqref{eqn:mtheta}.
The four approximations represent four different cases of approximation, that is, \eqref{eqn:rho} from training data to training data, \eqref{eqn:nutheta} from algorithm support to algorithm support, \eqref{eqn:ltilde} from training data to algorithm support, and \eqref{eqn:mtheta} from algorithm support to training data, respectively.
In our numerical experiments, we simply apply an isotropic Gaussian kernel
\begin{equation}\label{eqn:kernel}
\kappa(x,x') \propto \exp\left(\dfrac{\|x-x'\|_{\mathcal{X}}^2}{\epsilon^2}\right),
\end{equation}
while the variance $\epsilon^2$ can be different in each approximation.
Generally speaking, a smaller variance can give a more accurate estimate as long as the sample size is large enough.
However, larger variance will make the empirical risk $R_N$ better conditioned, which makes the iteration numerically stable for larger learning rates.
The constant coefficients in \eqref{eqn:kernel} and after are omitted because they will be canceled after normalization of $m_\theta$.

Because early stopping is implemented according to \eqref{eqn:rsn}, the variance in \eqref{eqn:ltilde} should be carefully chosen.
We apply a simple method to determine this variance.
By choosing a point $x_i$ in the training data, we can estimate the error $\|f_\theta(x_i) - f^*(x_i)\|^2_{\mathcal{Y}}$ using a method similar to \eqref{eqn:ltilde} as $l_{\theta,i}(x_i)$, where the training data $(x_i, f^*(x_i))$ is removed in this estimation.
Then, for some training data $x_i$ located near the algorithm support $\mathcal{S}(f_\theta)$. we can obtain the average estimation error as
\begin{equation}\label{eqn:remove}
\operatorname{mean} \left(l_{\theta,i}(x_i) - \|f_\theta(x_i), f^*(x_i)\|^2_{\mathcal{Y}}\right)^2.
\end{equation}
The variance of the kernel in \eqref{eqn:ltilde} is chosen to minimize this average estimation error \eqref{eqn:remove}.
Note that if our training data are unevenly distributed around each point in the algorithm support, a different kernel variance
may be applied for each point.
In each numerical example in this paper, we apply the same variance for each point for simplicity.

The kernel variance in \eqref{eqn:mtheta} indicates the neighborhood for each point in the algorithm support where we minimize the error of some training data.
A larger variance gives a larger region around the algorithm support in the task distribution where the surrogate model should have good predictive capabilities, so the surrogate model will be more robust for the algorithm.
In practice, it should be not smaller than the variance in \eqref{eqn:ltilde}.
Our method is not sensitive to the choices of variance in \eqref{eqn:rho} and \eqref{eqn:nutheta}.

\textbf{Determining reweighting coefficients.}
In \eqref{eqn:meansup}, we use $M$ to approach the supremum error \eqref{eqn:sup}.
The hyperparameter $\omega_0$ emphasizes the mean error on the algorithm support and gives a minimal weight of each point in the algorithm support,
A small $\omega_0$ or a large $M$ may make the empirical risk $R_N$ ill-conditioned, so each optimization step should be implemented more cautiously for numerical stability.
For example. with an overly large $M$, the support point with the largest error, namely $\tilde{x}$, will be emphasized in the reweighted empirical risk $L_N$.
We should slightly train the model to reduce $\|f_\theta(\tilde{x}) - f^*(\tilde{x})\|^2_{\mathcal{Y}}$ while the errors of other support points do not increase much.
If we implement training excessively, $\tilde{x}$ would be well approximated after training, while another support point, namely $\tilde{x}'$, would be of the largest error.
The training procedure may alternatively minimize $\tilde{x}$ and $\tilde{x}'$, which is highly ineffective.
Therefore, the learning rate and the epoch number should be small enough to ensure that the reweighted empirical risk $R_{N}$ does not oscillate violently during training.
Our method is not sensitive to $M$ or $\omega_0$, and we take $M=10$, $\omega_0=0.5$ in our numerical experiments.
To achieve a more significant improvement of task-specific supervised learning \cref{eqn:rp} over MSE \cref{eqn:p1}, we can apply a larger $M$ and a smaller $\omega_0$.

\textbf{Updating optimizers.}
Here we give a detailed description for updating the optimizer (Opt) in \cref{alg}.
At the beginning of our training, the optimizer is set as SGD with a small enough learning rate and a minimal training epoch number.
After each step of updating $m_\theta$, it should be checked whether the previous optimization is implemented successfully,
which can be monitored by $R_{\mathcal{S}, N}$ -- the same criterion for early stopping.
If $R_{\mathcal{S}, N}$ decreases, the previous optimization is believed to be well implemented, so a larger epoch number is accepted in the next iteration.
Although the gradient of $m_\theta$ is not evaluated, it can be estimated a posteriori by how much $m_\theta$ changes.
If the difference of $m_\theta$ between the last two iterations is small, then $m_\theta$ is near convergence and
a larger epoch number can also be accepted in the next iteration.
Otherwise, a smaller epoch number is applied for stability.
Moreover, if the optimization has been successful for many iterations, we can apply more powerful optimizers such as Adam \cite{kingma2014adam} to accelerate convergence.
In practice, the optimizer is changed to Adam with a much smaller learning rate if the training epoch number reaches a predetermined maximum value.
Furthermore, we add another criterion for Adam, in case that the optimization with improper reweighting coefficients leads to a bad model.
If $R_{\mathcal{S}, N}$ increases with the Adam optimizer, the model parameters are restored to the state in the previous iteration,
and the optimizer is reset as SGD with a smaller epoch number.
We apply these practical methods to accelerate convergence of training.

\textbf{Stopping criteria.}
The training procedure is terminated if $R_{\mathcal{S},N}$ remains almost unchanged for several iterations.
In our numerical experiments, the training is terminated if the standard error of $R_{\mathcal{S},N}$ in the last 10 iterations is smaller than $10^{-6}$.
If $R_{\mathcal{S},N}$ keeps increasing for many iterations, the training is also terminated because the training does not seem to converge.
Then a surrogate model is accepted based on early stopping strategy.

\bibliographystyle{siam}
\bibliography{references}